\documentclass[preprint,5p]{elsarticle}

\usepackage{lineno}
\usepackage[hyphens]{url}
\usepackage{hyperref}
\usepackage{chngcntr}
\usepackage{color,soul}

\journal{Neural Networks}

\biboptions{authoryear}

\usepackage{amsmath,amsfonts,amssymb}
\usepackage{siunitx}
\usepackage{pdfcomment}

\begin{document}

\begin{frontmatter}

\title{Effect of the initial configuration of weights\\ on the training and
function of artificial neural networks}

\author[deti,ue]{R.~J.~Jesus\corref{cor1}}\ead{ricardojesus@ua.pt}

\cortext[cor1]{Corresponding author}

\author[deti,it]{M.~L.~Antunes\corref{cor1}}\ead{mario.antunes@ua.pt}
\author[dfis]{R.~A.~da Costa\corref{cor1}}\ead{americo.costa@ua.pt}

\author[dfis]{S.~N.~Dorogovtsev}\ead{sdorogov@ua.pt}
\author[dfis]{J.~F.~F.~Mendes}\ead{jfmendes@ua.pt}

\author[deti,it]{R.~L.~Aguiar}\ead{ruilaa@ua.pt}

\address[deti]{Departamento de Eletrónica, Telecomunicações e Informática,
Universidade de Aveiro, Campus Universitário de Santiago, 3810--193 Aveiro,
Portugal}
\address[ue]{EPCC, The University of Edinburgh, Edinburgh, United Kingdom}
\address[it]{Instituto de Telecomunicações, Campus Universitário de Santiago,
3810--193 Aveiro, Portugal}
\address[dfis]{Departamento de Física $\&$ I3N, Universidade de Aveiro, Campus
Universitário de Santiago, 3810--193 Aveiro, Portugal}

\begin{abstract}
The function and performance of neural networks is largely determined by the evolution
of their weights and biases in the process of training, starting from the
initial configuration of these parameters to one of the local minima of the
loss function.
We perform the quantitative statistical characterization of the deviation of
the weights of two-hidden-layer ReLU networks of various sizes trained via
Stochastic Gradient Descent (SGD) from their initial random configuration.
We compare the evolution of the distribution function of this deviation with
the evolution of the loss during training. We observed that successful
training via SGD leaves the network in the close neighborhood of the initial
configuration of its weights. 
For each initial weight of a link we measured the distribution function of the
deviation from this value after training and found how the moments of this
distribution and its peak depend on the initial weight. We explored the
evolution of these deviations during training and observed an abrupt increase
within the overfitting region. 
This jump 
occurs simultaneously 
with a similarly abrupt
increase recorded in the evolution of the loss function. 
Our results suggest that 
SGD's ability to efficiently find local minima is restricted to the vicinity of the random initial configuration of weights.

\end{abstract}

\begin{keyword}
training\sep%
evolution of weights\sep%
deep learning\sep%
neural networks\sep%
artificial intelligence%
\end{keyword}

\end{frontmatter}

\section{Introduction}%
\label{s10}

Training of neural networks is based on the progressive correction of their
weights and biases (model parameters) performed by such algorithms as 
gradient descent which compare actual outputs with the desired ones
for a large set of input samples~\citep{lecun2015deep}. Consequently, the understanding
of the function of neural networks should be intrinsically based on the
detailed knowledge of the evolution of their weights in the process of training,
starting from their initial configuration. 
Recently,
\citet{li2018learning}
revealed that, during training, weights in neural networks only slightly deviate
from their initial values in most practical scenarios.
In this paper, we explore in detail the role of the initial configuration of
neural networks' weights in their training and function. We scan the
evolution of the weights of networks consisting of two ReLU hidden layers
trained on three different classification tasks with Stochastic Gradient Descent (SGD), and measure the
dependence of the distribution of deviations from an initial weight on
this initial value. 
In all of our experiments, we observe no inconsistencies on the results of the three tasks.

\begin{figure*}[!htb]
\centering
\includegraphics[width=.95\textwidth]{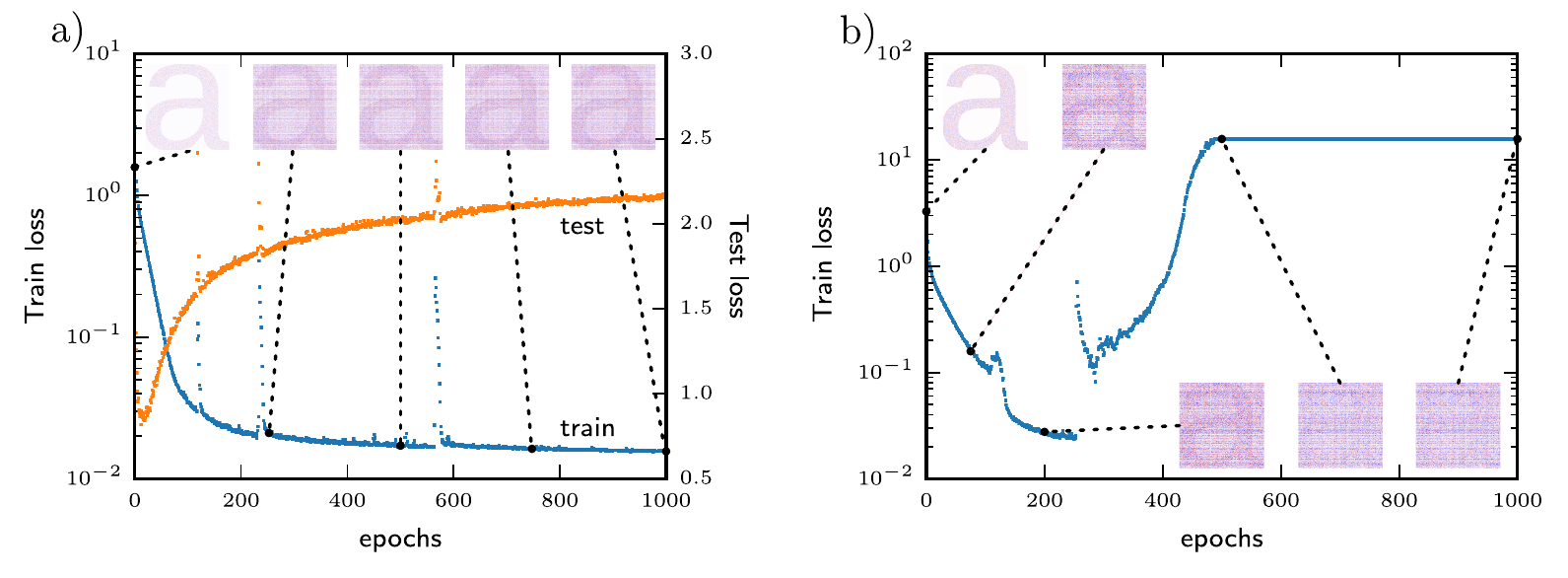}
\caption{Train and test loss of networks consisting of two equally sized hidden layers of nodes, trained on HASYv2.
Some of the weights connecting the two hidden layers
were initially set to zero so that the weight matrix of these layers resembles
the letter \texttt{a} (at initialization).
The evolution of the weight matrix is shown in the
subplots.
(a) Loss of a stable learner network with 512 nodes in each hidden layer.
(b) Training loss of an unstable network with 256 nodes in each hidden layer,
illustrating the effect of crossing over from trainability to untrainability
regimes on the network's weights (disappearance of the initialization mark).
A single curve is shown for clarity, but networks of the same width show the
same behavior.
Experiments with other symbols exhibit similar behavior.}%
\label{f50090}
\end{figure*}

\begin{figure*}[!tbh]
\centering
\includegraphics[width=\textwidth]{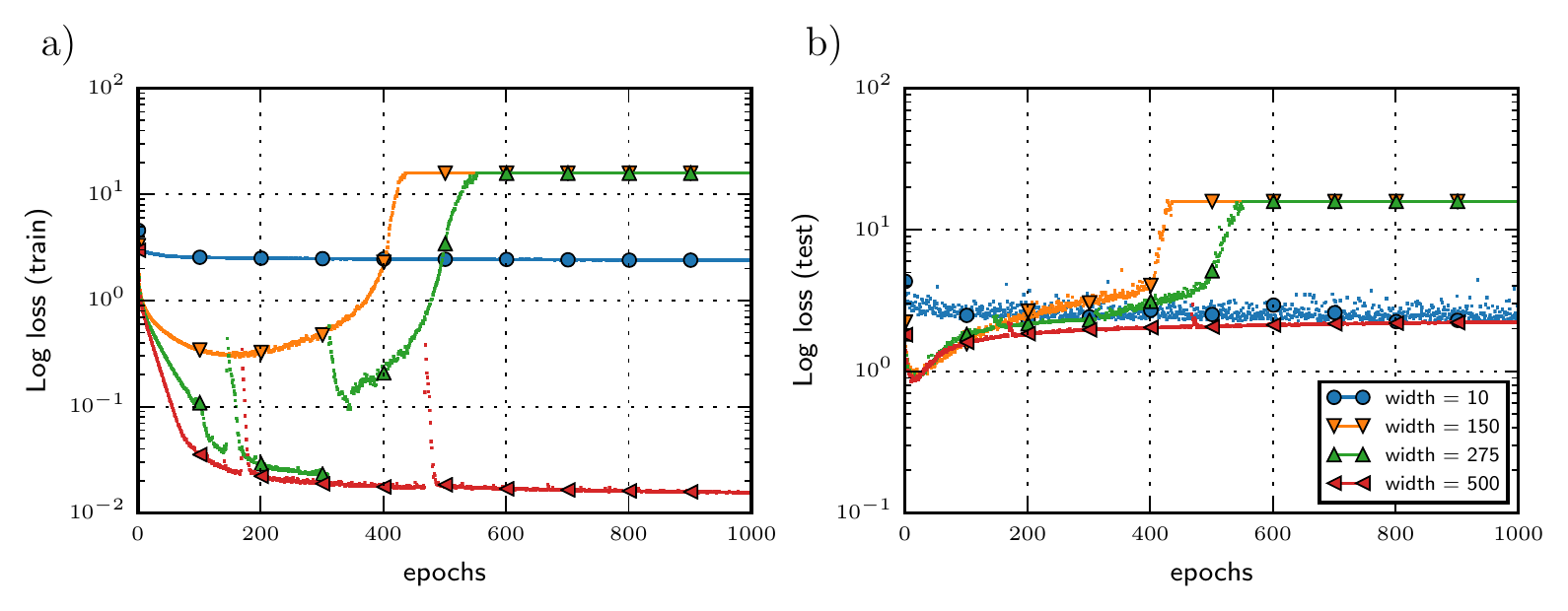}
\includegraphics[width=\textwidth]{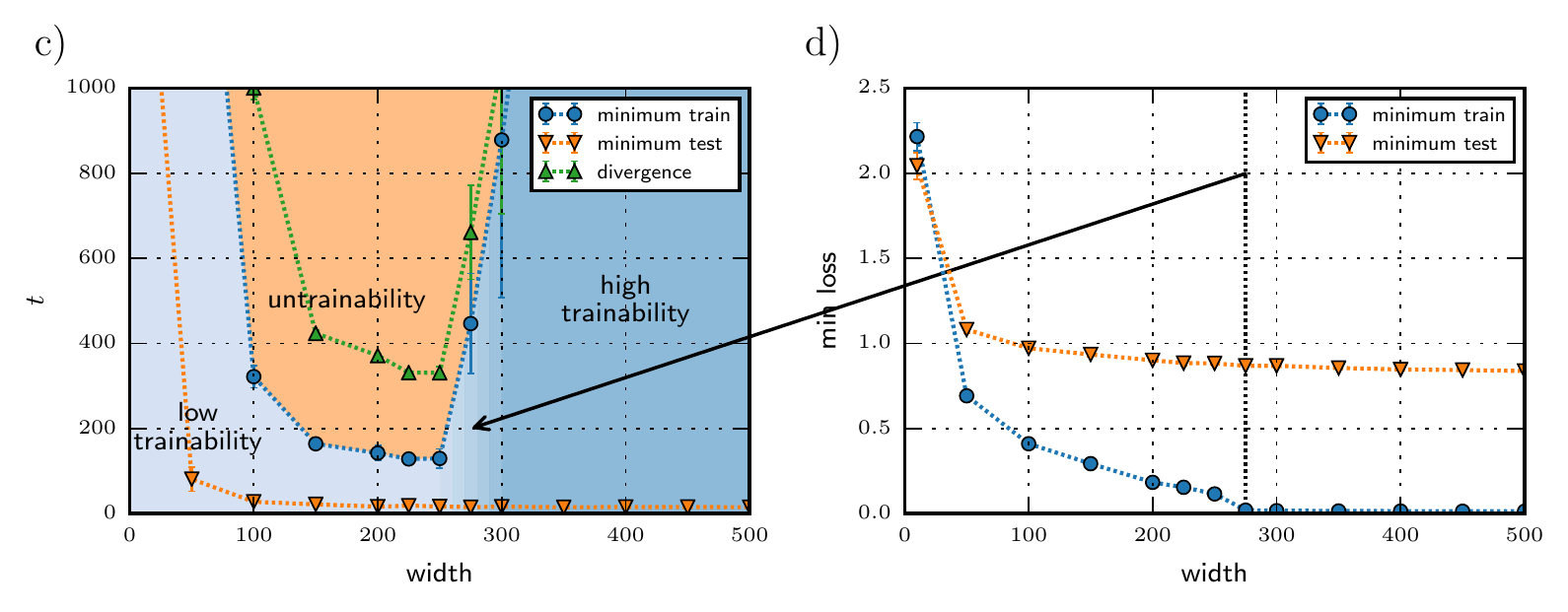}
\caption{The regimes of a neural network over the course of its training:
Evolution of (a) train and (b) test loss functions of networks of
various widths. Panels (a) and (b) show single typical training runs of the full set which we explored. (c) Average times (in epochs) taken by the networks to
reach the minima of the train loss and of the test loss functions, and to diverge (i.e., reach the plateau of the train loss).
For each network width, we calculate the averages and standard deviations of these times
(represented by error bars)
over ten independent runs trained on HASYv2.
(d) Average values of minimum loss at the train and
test sets reached during individual runs (different runs reach their minima at different times). 
These averages were measured over the same ten independent realizations as in panel(c).
}%
\label{f00099}
\end{figure*}

By considering networks of different sizes we observe that, in order to reach
an
arbitrarily chosen loss value, the weights of larger networks tend to deviate
less from their initial values (on average) than the weights of smaller
networks.
This suggests that larger networks tend to
converge to minima which are closer to their initialization. 
On the other hand, we observe that for a certain range of network sizes the
deviations from initial weights abruptly increase at some moment during their
training within the
overfitting regime, see Fig.~\ref{f50090}. We find that this sharp increase closely correlates with
the crossover between two regimes of the network---trainability and
untrainability---occurring in the course of the training.
Finally, we measure the dependence of the time at which these crossovers
happen on the network size and build a diagram showing
the network's training regime
for each network width and training time.
This diagram, Fig.~\ref{f00099}~(c), resembles a phase diagram, although the variable $t$, the training time, is not a control parameter, but it is rather the measure of the duration of the `relaxation' process that SGD training represents.

One may speak about a phase transition in these systems only in respect of their stationary state, that is, the regime in which they finally end up, after being trained for a very long (infinite) time.
Figure~\ref{f00099}~(c) shows three characteristic instants (times) of training for each network width: (i) the time at which the minimum of the test loss occurs, (ii) the time of the minimum of the train loss, (iii) the time at which loss abruptly increases (`diverges'). 
Each of these times differ for different runs, and, for some widths, these fluctuations are strong or they even diverge. 
The points in this plot are the average values over ten independent runs. 
In the error bars we show the scale of the fluctuations between different runs.
Notice that the times (ii) and (iii) approach infinity as we approach the threshold of about 300 nodes from below (which is specific to the networks' architecture and dataset).
Therefore, wide networks ($ \gtrsim 300$ nodes in each hidden layer) never cross over to the untrainability regime; wide networks should to stabilize in the trainability regime as $t \to \infty$.
The untrainability region of the diagram exists only for widths smaller than the threshold, 
which is in the neighborhood of 300 nodes.
Networks with such widths initially demonstrate a consistent decrease of the train loss.
But, at some later moment, during the training process, the systems abruptly cross over from the trainability regime, with small deviations of weights from their initial
values and decreasing train loss, to the untrainability regime, with large loss and large deviations of weights.

By gradually reducing the width, and looking at the trainability regime in the limit of infinite training time,
 we find a phase transition from trainable to untrainable networks.
In the diagram of Fig.~\ref{f00099}~(c), this transition corresponds to an horizontal line at $t=\infty$, or, equivalently, the projection of the diagram on the horizontal axis (notice that the border between regimes is concave).
Note that we use the term trainability/untrainability referring to the regimes of the training process, in which loss and deviations of weights are, respectively,  small/large.
We reserve the terms trainable-untrainable transition to refer to the capability of a network to keep a low train loss after infinite training, which depends essentially on the network's architecture.

Our paper is organized as follows. In Sec.~\ref{s20} we summarize some
background topics on neural networks' initializations, and review a series of
recent papers pertaining to ours. 
Section~\ref{s25} presents the experimental settings and datasets used in this
paper.
In Sec.~\ref{s30} we explore the shape of the distribution of the deviations
of weights from their initial values and its dependence on the initial
weights.
We continue these studies in Sec.~\ref{s40} by experimenting with networks of
different widths and find that, whenever a network's training is successful,
the network does not travel far from its initial configuration. Finally,
Sec.~\ref{s60} provides concluding remarks and points out directions for
future research.

\section{Background and related work}%
\label{s20}

\subsection{Previous works}


It is widely known that a neural network's initialization is instrumental in
its training~\citep{lecun1998efficient, yam2000weight,
glorot2010understanding, he2015delving}. The works of
\citet{glorot2010understanding}, \citet{chapelle2011improved} and
\citet{krizhevsky2012imagenet}, for instance, showed that deep networks
initialized with random weights and optimized with methods as simple as
Stochastic Gradient Descent could, surprisingly, be trained successfully. In
fact, by combining momentum with a well-chosen random initialization strategy,
\citet{sutskever2013importance} managed to achieve performance comparable to
that of Hessian-free methods.

There are many methods to randomly initialize a network. Usually, they consist
of drawing the initial weights of the network from uniform or Gaussian
distributions centered at zero, and setting the biases to zero or some other
small constant.
While the choice of the distribution (uniform or Gaussian) does not seem to be
particularly important~\citep{goodfellow2016deep}, the scale of the
distribution from which the initial weights are drawn does. The most common
initialization strategies --- those of \citet{glorot2010understanding},
\citet{he2015delving}, and \citet{lecun1998efficient} --- define rules based
on the network's architecture for choosing the variance that the distribution
of initial weights should have. These and other frequently used initialization
strategies are mainly heuristic, seeking to achieve some desired properties at
least during the first few iterations of training. However, it is generally
unclear which properties are kept during training or how they
vanish~\citep[Sec.~8.4]{goodfellow2016deep}. Moreover, it is also not clear
why some initializations are better from the point of view of optimization
(i.e., achieve lower training loss), but are simultaneously worse from the
point of view of generalization.

\citet{frankle2018lottery} recently observed that randomly initialized dense
neural networks typically contain subnetworks (called winning tickets) that
are capable of matching the test accuracy of the original network when trained
for the same amount of time in isolation. Based on this observation they
formulate the Lottery Ticket Hypothesis, which essentially states that this
effect is general and manifests with high probability in this kind of
networks. Notably, these subnetworks are part of the network's initialization,
as opposed to an organization that emerges throughout training.
The subsequent works of \citet{zhou2019deconstructing} and
\citet{ramanujan2019s}
corroborate the Lottery Ticket Hypothesis and propose that winning tickets may
not even require training to achieve quality comparable to that of the trained
networks.

In their recent paper, \citet{li2018learning} established that
two-layer over-parameterized ReLU networks, optimized with SGD on data drawn
from a mixture of well-separated distributions, probably converge to a minimum
close to their random initializations. Around the same time,
\citet{jacot2018neural} proposed the neural tangent kernel (NTK), a kernel
that characterizes the dynamics of the training process of neural networks in
the so-called infinite-width limit. These works instigated a series of
theoretical breakthroughs, such as the proof that SGD can find global minima
under conditions commonly found in practice (e.g., over-parameterization)
\citep{du2019gradient, du2019gradient2, zhu2019learning, zhu2019convergence,
zhu2019on, oymak2019overparameterized, 9081945, zou2020gradient}, and that, in the
infinite-width limit, neural networks remain in an $O\left(1/\sqrt{n}\right)$
neighborhood of their random initialization ($n$ being the width of the hidden
layers) \citep{arora2019fine,NEURIPS2019_dbc4d84b}. \citet{lee2019wide} make a similar claim about
the distance a network may deviates from its linearized version.
\citet{chizat2019lazy}, however, argue that such wide networks operate in a
regime of ``lazy training'' that appears to be incompatible with the many
successes neural networks are known for in difficult, high dimensional tasks.

\subsection{Our contribution}

\begin{figure}[!htb]
\centering
\includegraphics[width=.5\textwidth]{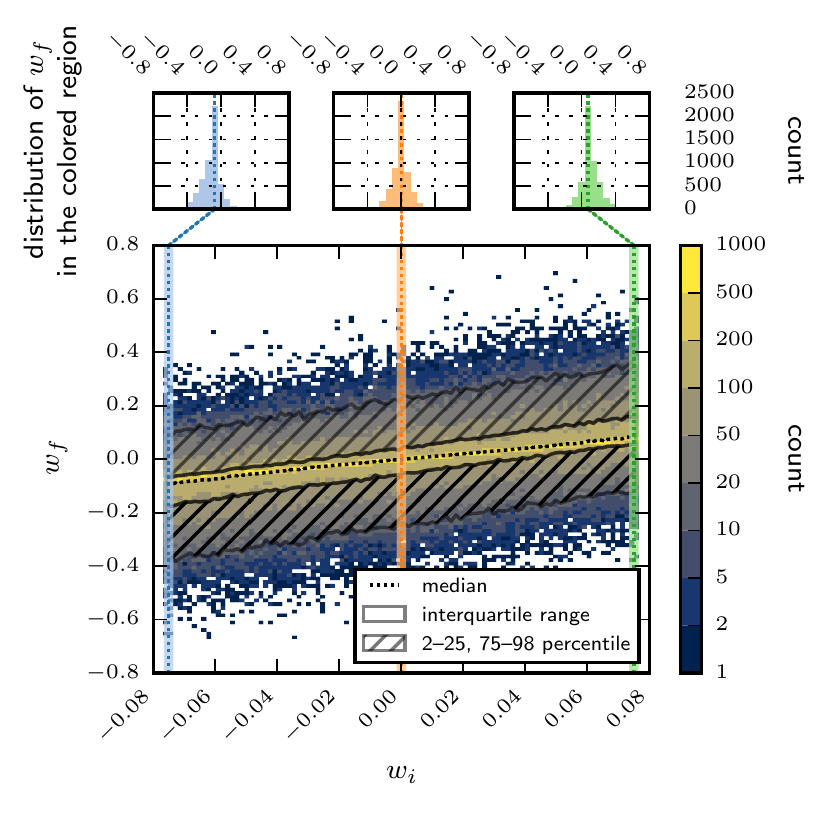}
\caption{Distribution of the final values of the weights of the
network
of Fig.~\ref{f50090}~(a), trained for 1000 epochs on HASYv2, as function of their initial
value. The peak of the distribution is at $w_{f} = w_{i}$, which is extremely
close to the median. The skewness of the distribution for large absolute
values of $w_{i}$ is evidenced in the histograms at the top.}%
\label{f50125}
\end{figure}

From distinct perspectives, these previous works have shown that, in high over-parametrized networks, the training process consists in a fine-tuning of the initial configuration of weights, adjusting significantly just a small portion of them (the ones belonging to the winning tickets).
Furthermore, \citet{frankle2020} recently showed that the winning ticket's weights are highly correlated with each other.
The effects of having a winning ticket are visible in Fig.~\ref{f50125}. 
This figure demonstrates the highly structured correlation between the weights before and after training, including that most of them are left essentially unchanged by the training.
Our results suggest that the role of the over-parametrized initial configuration is actually decisive in successful training: 
when we reduce the level of over-parametrization to a point where the initial configuration stops containing such winning tickets, the network becomes untrainable by SGD.

The previous investigations on the role of the initial weights configuration focus on large networks, in which, as our results also show, the persistence of the initial configuration is more noticeable.
In contrast, we explore a wide range of network sizes from untrainable to trainable by varying the number of units in the hidden layers.
This approach allows us to explore the limits of trainability, and characterize the trainable-untrainable network transition that occurs at a certain threshold of the width of hidden layers, see Fig.~\ref{f00099}.

A few recent works \citep{lu2017expressive,li2018benefit} indicated the existence of `phase transitions' from narrow to wide
networks associated to qualitative changes in the set of loss minima in the configuration space.
These results resonate with ours, although neither relations to trainability nor the role of the initial configuration of weights were explored.

On one hand, we observe that, when the networks are trainable (large networks), they always converge to a minima in the vicinity of the initial weight configuration.
On the other hand, when the network is untrainable (small networks) the weight configuration drifts away from the initial one. 
There is an intermediate size range for which the networks train reasonably well for a while, decreasing the loss consistently, but, later in the overfitting region, their loss abruptly increases dramatically (due to overshooting).
Past this point of divergence, the loss can no longer be reduced by more training.
The behavior of these ultimately untrainable networks further emphasizes the connection between trainability (ability to reduce train loss) and proximity to the initial configuration: the distance to the initial configuration remains small in the first stage of training, while the loss is reduced, and later increases abruptly, simultaneously with the loss.

We hypothesize that networks initialized with random weights and trained with SGD can \textit{only} find good minima in the vicinity of the initial configuration of weights. 
The training process has no ability to effectively explore more than a relatively small region of the configuration space around the initial point.

\section{Datasets and experimental settings}%
\label{s25}

\begin{figure}[!htb]
    \centering
    \includegraphics[width=.45\textwidth]{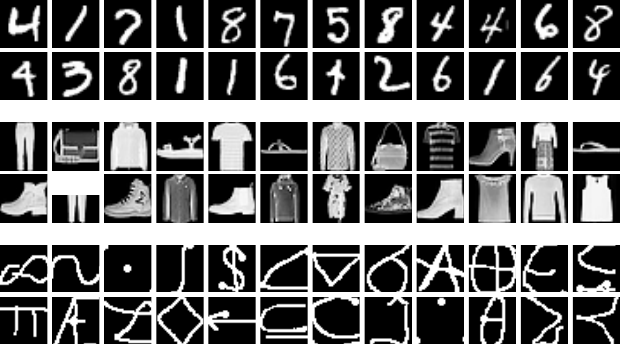}
    \caption{Samples of the datasets used in our experiments. Top: MNIST\@. Middle:
    Fashion MNIST\@. Bottom: HASYv2 (colors reversed).}%
    \label{f00005}
\end{figure}

Throughout this paper we use three datasets to train our networks: MNIST,
Fashion MNIST, and HASYv2. Figure~\ref{f00005} displays samples of them.

MNIST\footnote{\url{http://yann.lecun.com/exdb/mnist/}}~\citep{lecun1998gradient}
is a database of gray-scale handwritten digits. It consists of \num{60000}
training and \num{10000} test images of size $28{\times}28$, each showing one
of the numerals 0 to 9. It was chosen due to its popularity and widespread
familiarity.

Fashion
MNIST\footnote{\url{https://github.com/zalandoresearch/fashion-mnist}}~\citep{xiao2017fashion}
is a dataset intended to be a drop-in replacement for the original MNIST
dataset for machine learning experiments. It features 10 classes of clothing
categories (e.g., coat, shirt, etc) and it is otherwise very similar to
MNIST\@.  It also consists of $28{\times}28$ gray-scale images, \num{60000}
samples for training, and \num{10000} for testing.

HASYv2\footnote{\url{https://github.com/MartinThoma/HASY}}~\citep{thoma2017hasyv2}
is a dataset of $32{\times}32$ binary images of handwritten symbols (mostly
\LaTeX{} symbols, such as $\alpha$, $\sigma$, $\int$, etc). It mainly
differentiates from the previous two datasets in that it has many more classes
(369) and is much larger (containing around \num{150000} train and \num{17000}
test images).

We trained feedforward neural networks with two layers of hidden nodes (three layers of links), each containing
between 10 to 1000 units. The two hidden layers employ the ReLU activation
function, whereas the output layer employs softmax. This architecture is
largely based on the multilayer perceptron created by Keras for
MNIST\footnote{\url{https://raw.githubusercontent.com/keras-team/keras/master/examples/mnist_mlp.py}}.
Unless otherwise stated, the biases of the networks are initialized at zero and
the weights are initialized with Glorot's uniform
initialization~\citep{glorot2010understanding}:
\begin{equation}
    w_{ij} \sim U\left(-\frac{\sqrt{6}}{\sqrt{m+n}},\,
    \frac{\sqrt{6}}{\sqrt{m+n}}\right),
\label{e10}
\end{equation}
where $U(-x, x)$ is the uniform distribution in the interval $(-x, x)$, and
$m$ and $n$ are the number of units of the layers that weight $w_{ij}$
connects. In some of our experiments we apply various masks to these uniformly
distributed weights, setting to zero all weights $w_{ij}$ uncovered by a mask
(see Fig.~\ref{f50090}). The loss function minimized is the categorical
cross-entropy, i.e.,
\begin{equation}
    L = -\sum_{i=1}^{C} y_{i} \ln{o_{i}},
\end{equation}
where $C$ is the number of output classes, $y_{i} \in \{0,1\}$ the $i$-th
target output, and $o_{i}$ the $i$-th output of the network. The neural
networks were optimized with Stochastic Gradient Descent with a learning rate
of 0.1 and in mini-batches of size 128. The networks were defined and trained
in Keras~\citep{chollet2015keras} using its
TensorFlow~\citep{tensorflow2015whitepaper} back-end.

We typically trained networks for very long periods (1000 epochs), and
consequently for most of their training the networks were in the overfitting
regime.
However, since we are studying the training process of these network
and making no claims concerning the networks' ability to generalize on
different data, overfitting does not affect our conclusions. In fact, our
results are usually even stronger prior to overfitting. For similar reasons, we
will be considering only the loss function of the networks (and not other
metrics such as their accuracy), since it is the loss function that the
networks are optimizing.

\section{Statistics of deviations of weights from initial values}%
\label{s30}

To illustrate the reduced scale of the deviations of weights during the training, let us mark a network's initial configuration of weights
using a mask in the shape of a letter, and observe how the marking evolves as the network is
trained\footnote{The letters are stamped to a network's initial configuration
of weights by creating a bitmap of the same shape as the matrix of weights of
the layer being marked, rasterizing the letter to the bitmap, and using the
resulting binary mask to set to zero the weights laying outside the mark's area.}.
Naturally, if the mark is still visible after the weights undergo a
large number of updates and the networks converge, one may conclude that the training process does not shift the weights of a network far
from their initial state.

Figure~\ref{f50090}~(a) shows typical results of training a large network whose initial
configuration is marked with the letter \texttt{a}. One can see that the letter is clearly visible after training for
as many as 1000 epochs. In fact, one observes the
initial mark during all of the network's training, without any sign that it
will disappear.
Even more surprisingly, these marks do not affect the quality of training.
Independently of the shape marked (or whether there is a mark or not) the
network trains to approximately the same loss across different realizations of initial weights.
This demonstrates that
randomly initialized networks preserve features of their initial configuration
along their whole training---features that are ultimately transferred into the
networks' final applications.

Figure~\ref{f50090}~(b) demonstrates an opposite effect for midsize networks that
cross over between the regimes of trainability and untrainability. As it
illustrates, the initial configuration of weights of these unstable networks tend to
be progressively lost, suffering the largest changes when the networks diverge
(loss function sharply increases at some moment).

By inspecting the distribution of the final (i.e., last, after training) values of the weights of the network of Fig.~\ref{f50090}~(a)
versus their initial values, portrayed in Fig.~\ref{f50125}, we see that
weights that start with larger absolute values are more likely to suffer
larger updates (in the direction that their sign points to). This tendency can
be observed on the plot by the tilt of the interquartile range (yellow region
in the middle) with respect to the median (dotted line). 
The figure demonstrates
that initially large weights in absolute value have a tendency to
become even larger, keeping their original sign;
it
also shows the
maximum concentration of weights near the line $w_{f} = w_{i}$, indicating
that most weights either
change very little or
nothing at all throughout training.

The skewness in the distribution of the final
weights
can be explained by 
the randomness 
 of the initial configuration,
which initializes certain groups of
wights with more appropriate values than the others,
making few weights better for certain features of the dataset that is being
used for training the network. 
This subset of
weights 
does
not need to be particularly
good, but as long as it provides slightly better or more consistent 
outputs than the rest of the weights, the learning process will favor
their training, improving them further in comparison to the remaining weights.
Over the course of many epochs, the small preference that the learning
algorithm keeps giving them adds up and cause these weights to become
the best recognizers for the features that they initially, by chance, happened
to be better at. This effect has several bearing with, for instance, the
real-life effect of the month of birth in sports~\citep{helsen2005relative}.
In this hypothesis, it is highly likely that weights with larger initial
values are more prone to be deemed more important by the learning algorithm,
which will try to amplify their `signal'.

\section{Evolution of deviations of weights and trainability}%
\label{s40}

One may understand the
compatibility between the
success of training and the fine-tuning process observed in
Sec.~\ref{s30}, during which a large fraction of the weights of a network
suffer very tiny updates (and many are not even changed at all), in the
following way.  We suggest that the neural networks typically trained are so
over-parameterized that, when initialized at random, their initial
configuration has a high probability of being close to a proper minimum (i.e., a global minimum where the training loss approaches zero). 
Hence, to reach such a minimum, the network needs to adjust its weights only
slightly, which causes its final configuration of weights to have strong
traces of the initial configuration (in agreement with our observations).

\begin{figure*}[!tbh]
\centering
\includegraphics[width=\textwidth]{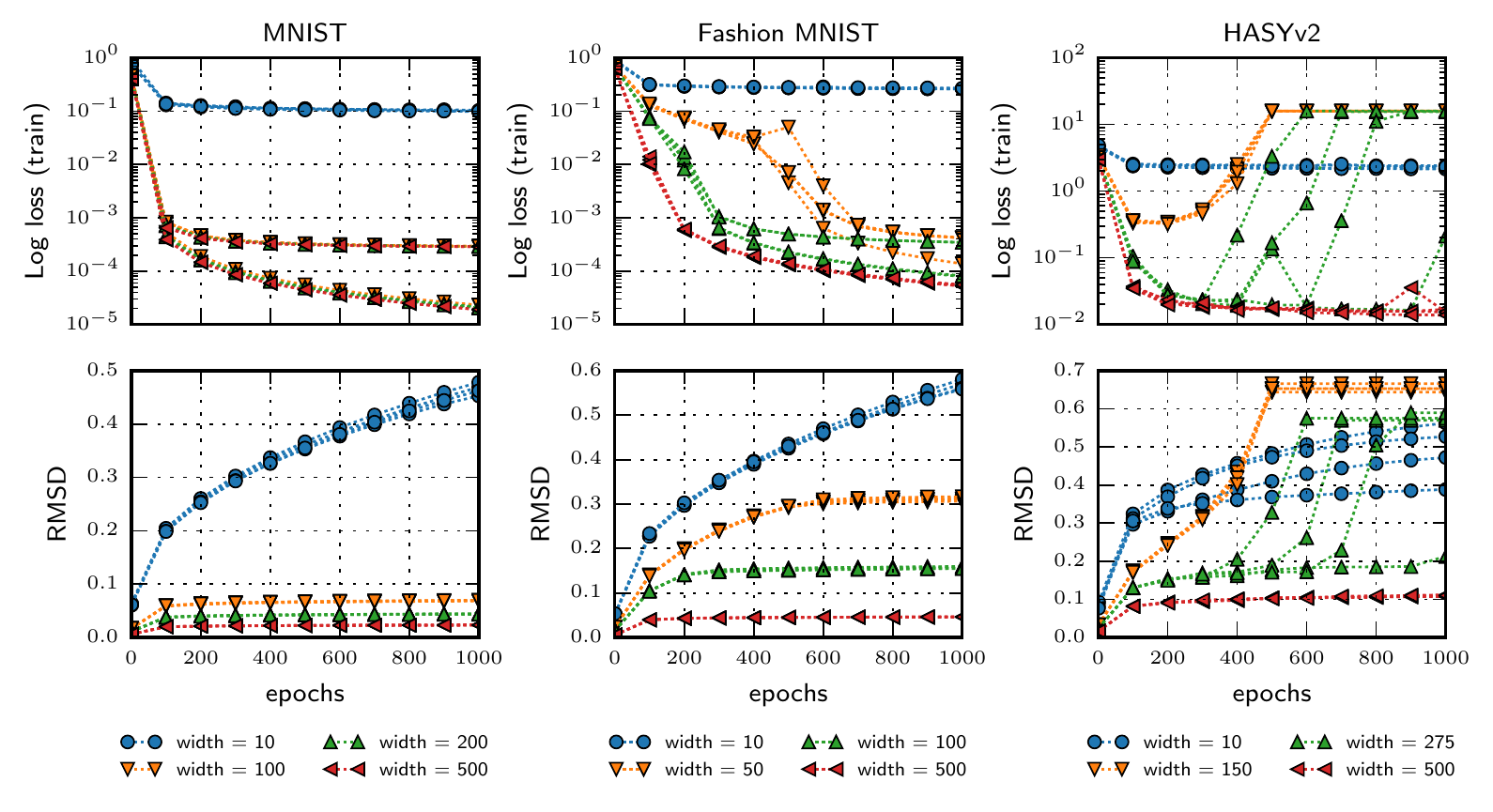}
\caption{Top: Temporal evolution of the loss function of networks of various
widths. Bottom: Evolution of the root mean square deviation (RMSD) between the
initial configuration of weights of a network and its current configuration.
From left to right: networks trained on the MNIST, Fashion MNIST, and HASYv2
datasets. Five independent test runs are plotted (individually) for each value
of width and each dataset.}%
\label{f00020}
\end{figure*}

This hypothesis raises the question of what happens when we train networks
that have a small number of parameters. At some point, do they simply start to
train worse? Or do they stop training at all? It turns out that, during the
course of their training, neural networks do cross over between two
regimes---trainability and untrainability. 
The trainability region may be further split into two distinct regimes:
a regime of \textit{high trainability} where training drives the networks towards global minima (with zero training loss), and a regime of \textit{low trainability} where the networks converge to sub-optimal minima of significantly higher loss.
Only high trainability allows a network to train successfully (i.e., to be trainable),
since in the remaining two regimes, of untrainability and low trainability,
either the networks do not learn at all, or they learn but very poorly.
Figure~\ref{f00099} illustrates these three regimes.

The phase diagram in this figure (the bottom left panel) was obtained by
analysing the temporal evolution of the train and test loss functions.  It
demonstrates the existence of three different classes of networks: weak,
strong, and unstable learners.
Weak learners are small networks that, over the course of their training, do
not tend to diverge, but only train to poor minima. They mostly operate in a
regime of low trainability, since they can be trained, but only to ill-suited
solutions. On the other spectrum of trainability are large networks.  These
are strong learners, as they train to very small loss values and they are not
prone to diverge (they operate mostly in the regime of high trainability). In
between these two classes are unstable learners, which are midsize networks
that train to progressively better minima (as their size increases), but that
over the course of their training are likely to diverge and become untrainable
(i.e., they tend to show a crossover from the regime of trainability to the
one of untrainability).

Remarkably, we observe the different regimes of operation of a network not
only from the behavior of its loss function, but also from how far it travels
from its initial configuration of weights. 
We have already demonstrated
qualitatively in Fig.~\ref{f50090} how the mark of the initial configuration of weights of
a network persists in large networks (i.e., strong learners that over the
course of their training were always on the regime of high trainability), and vanishes for midsize networks that
ultimately cross over to the regime of untrainability. 
In~\ref{appendixA} we supply detailed description of the evolution of the statistics of weights during the training of the networks used to draw Fig.~\ref{f00099}.
Figures~\ref{f01131}  and~\ref{f01126} show that, as the network width is reduced, the highly structured correlation between initial and final weight, illustrated by the coincidence of the median with the line $w_f=w_i$ (see Fig.~\ref{f50125}), remains in effect in all layers of weights until the trainability threshold. Below that point the structure of the correlations eventually breaks down, given enough training time.

To quantitatively describe how distant a network gets from its initial
configuration of weights we consider the root mean square deviation (RMSD) of
its system of weights at time $t$ with respect to its initial configuration,
i.e.,
\begin{equation}
    \text{RMSD}(t) \equiv \sqrt{\frac{1}{m} \sum_{j=1}^{m} {\left[w_{j}(t) -
    w_{j}(0)\right]}^2},
\end{equation}
where $m$ is the number of weights of the network (which depends on its
width), and $w_{j}(t)$ is the weight of edge $j$ at time $t$.

Figure~\ref{f00020} plots, for three different datasets, the evolution of the
loss function of networks of various widths alongside the deviation of
their configuration of weights from its initial state. These plots evidence
the existence of a link between the distance a network travels away from its
initialization and the regime in which it is operating, which we describe
below.

For all the datasets considered, the blue circles ($\bullet$) show the
training of networks that are weak learners---hence, they only achieve very
high losses and are continuously operating in a regime of low trainability.
These networks experience very large deviations on their configuration of
weights, getting further and further away from their initial state. In
contrast, the red left triangles ($\blacktriangleleft$) show the training of
large networks that are strong learners (in fact, for MNIST all the networks
marked with triangles are strong learners; in our experiments we could not
identify unstable learners on this dataset). These networks are always
operating in the regime of high trainability, and over the course of their
training they deviate very slightly from their initial configuration
[compare with the results of \citet{li2018learning}].
Finally,
for the Fashion MNIST and HASYv2 datasets, orange down ($\blacktriangledown$)
and green up ($\blacktriangle$) triangles show unstable networks of different
widths (the former being smaller than the latter). While on the regime of
trainability, these networks deviate much further from their initial
configuration than strong learners (but less than weak learners). However, as
they diverge and cross over into the untrainability regime (which could
only be observed on networks training with the HASYv2 dataset), the RMSD suffers a
sharp increase and reaches a plateau.
These observations highlight the persistent coupling between the 
network's trainability (measured as train loss) and the distance it travels 
away from the initial configuration (measured as RMSD), as well as their 
dependence of the networks's width.

\begin{figure*}[!tbh]
\centering
\includegraphics[width=\textwidth]{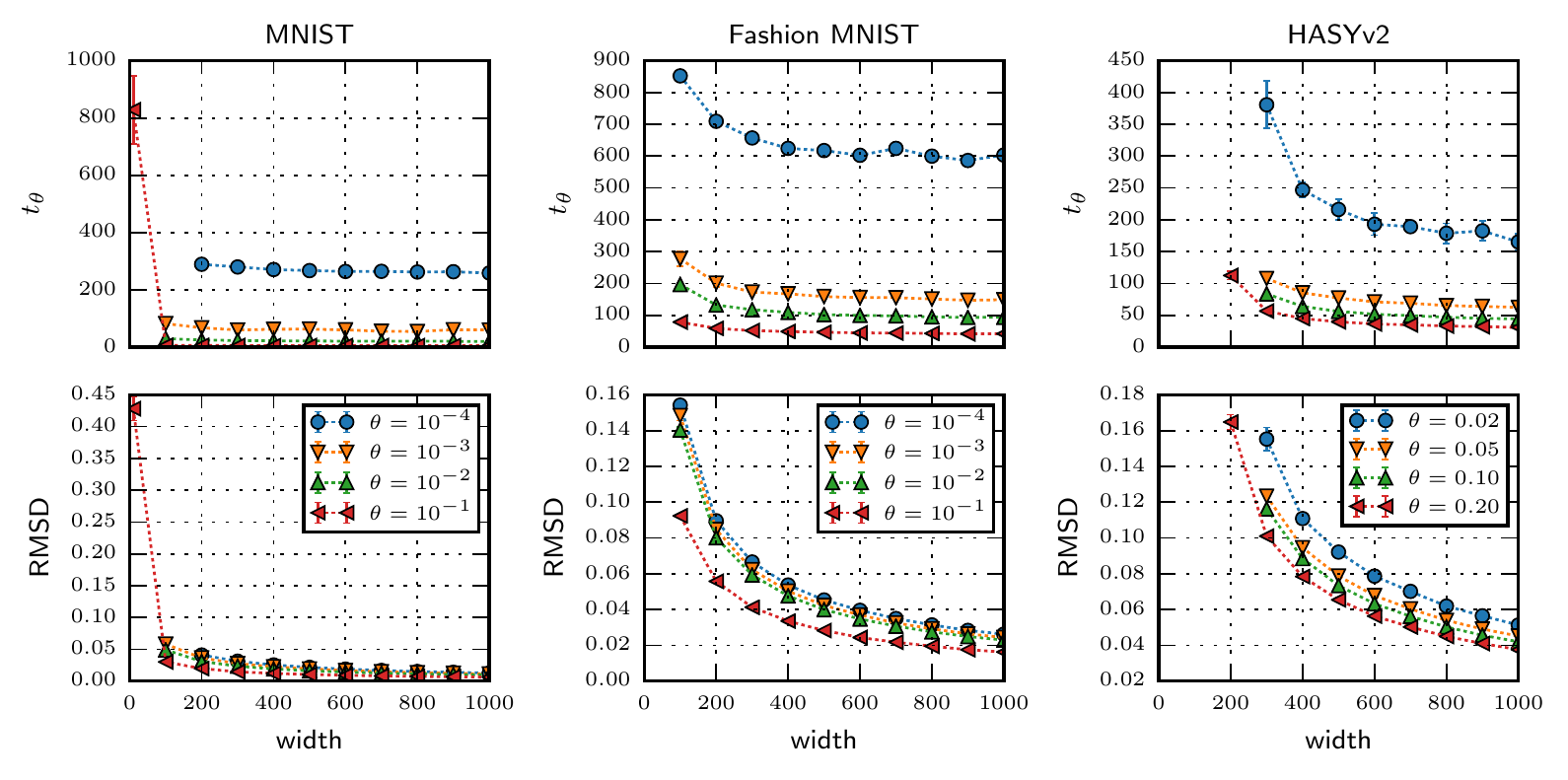}
\caption{Top: Time $t_{\theta}$ (in epochs) at which the networks first reach
a given value $\theta$. Bottom: Root mean square deviation (RMSD) between the
initial configuration of weights of a network and its configuration of weights
at time $t_{\theta}$. From left to right: networks trained on the MNIST,
Fashion MNIST, and HASYv2 datasets. Each point in the figure is the average of
five independent test runs. 
The absence of a point in a plot indicates that the network does not reach this loss during the entire period of training.
}%
\label{f00010}
\end{figure*}

To complete the description of the behavior of these networks on the different
regimes, Fig.~\ref{f00010} plots, for networks of different widths, the time
at which they reach a loss below a certain value $\theta$, and the RMSD
between their configuration of weights at that time and the initial one. It
shows that networks that are small and are operating under the low
trainability regime fail to reach even moderate losses (e.g., on Fashion
MNIST, no network of width 10 reaches a loss of 0.1, whereas networks of width
100 reach losses that are three orders of magnitude smaller). Moreover, even
when they reach these loss values, they take a significantly larger time to do
so, as the plots for MNIST demonstrate.
Finally, the figure also shows that, as the networks grow in size, the
displacement each weight has to undergo to allow the network to reach a
particular loss decreases, meaning that the networks are progressively
converging to minima that are closer to their initialization. We can treat
this displacement as a measure of the work the optimization algorithm performs
during the training of a network to make it reach
a particular quality (i.e., value of loss). Then one can say that using larger
networks eases training by decreasing the average work the optimizer has to
spend with each weight.

\section{Conclusions}%
\label{s60}

In this paper we explored the effects of the initial configuration of weights
on the training and function of neural networks. 
We performed the statistical characterization of the deviation of the weights of two-hidden-layer
networks of various sizes trained via Stochastic Gradient Descent from their initial random configuration. 
We observed that the initial configuration of weights typically leaves recognizable traces on the final configuration after training, which confirms that the learning
process is based on fine-tuning the weights of the network. 
We investigated 
the conditions under which a network travels far
from its initial configurations. We observed that a neural network learns in
one of two major regimes: trainability and untrainability. Moreover, its size
(number of parameters) largely determines the quality of its training process
and its chance to enter the untrainability regime.

We compared the evolution of the distribution function of this deviation with the evolution of the loss during training and observed that
over the course of training a network travels considerably
far away from its initial configuration of weights only if it is either (i) a
poor learner (which means that it never reaches a good minimum) or (ii) when
it crosses over from trainability to untrainability regimes. In the alternative
(good) case, in which the network is a strong learner and it does not become
untrainable, the network always converges to the neighborhood of its initial
configuration (keeping extensive traces of its initialization); in all of our 
experiments, we never observed a network converging to a good minimum
outside the vicinity of the initial configuration.

Note that most of our results 
are for
times when overfitting is
already taking place. At shorter times, the deviations of weights from their
initial values are even smaller, and our conclusions remain valid.

We based our conclusions on the analysis of the loss function of the networks
we trained, since this was the actual function that the networks were
optimizing. However, equivalent results can be obtained by using the accuracy
or other similar metric.

We intend to verify the generality of our results in neural networks of
different architectures and with datasets representative of more challenging
tasks. 
We also intend to find out
whether our results hold while varying the depth of
the networks.

\section*{Acknowledgements}

This work was developed within the scope of the project i3N, 
UIDB/50025/2020 and UIDP/50025/2020,
financed by national funds through the FCT/MEC.
RAC acknowledges the FCT Grant No. CEECIND/04697/2017.
The work was also supported under the framework of the New Generation Architecture project inside IT-Aveiro, by FCT\slash MCTES through national funds and when applicable co-funded EU funds under the project UIDB\slash 50008\slash2020-UIDP\slash 50008\slash 2020, under the project PTDC\slash EEI-TEL\slash 30685\slash 2017 and by the Integrated Programme of SR\&TD ``SOCA'' (Ref. CENTRO-01-0145-FEDER-000010), co-funded by Centro 2020 program, Portugal 2020, European Union, through the European Regional Development Fund.

\appendix

\section{Statistics of weights across the trainable-untrainable transition}
\label{appendixA}

\counterwithin{figure}{section}
\setcounter{figure}{0}
\renewcommand\thefigure{A.\arabic{figure}}

\begin{figure*}[!tbh]
\centering
\includegraphics[width=\textwidth]{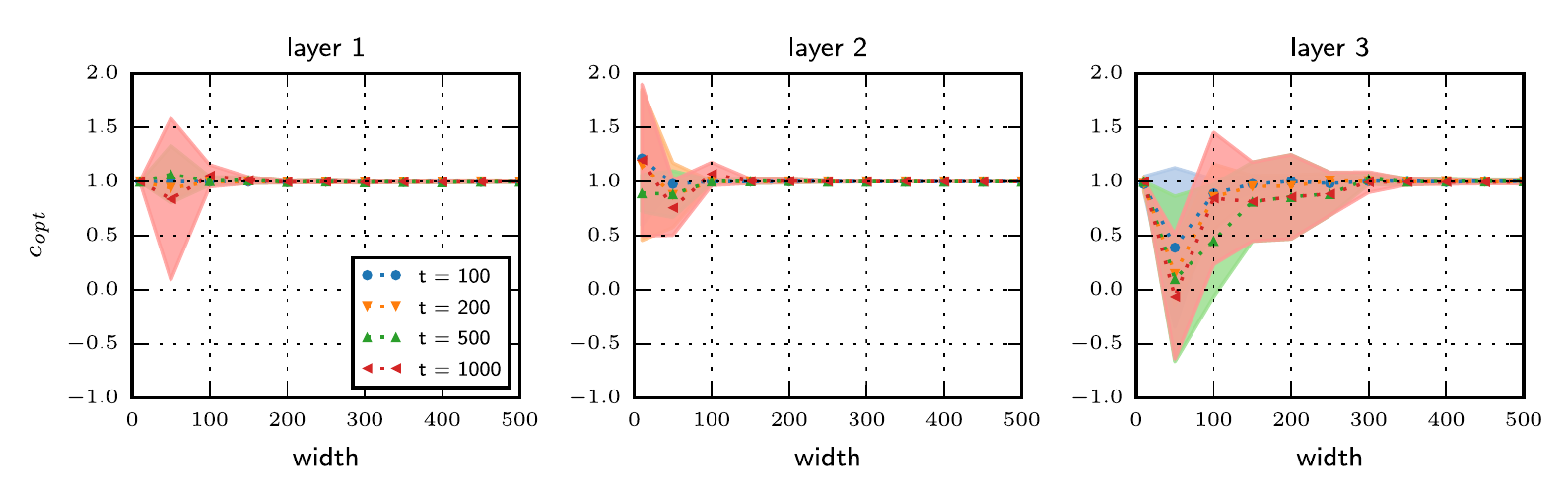}
\caption{
Fitting of the line across the maxima of the distribution of final weights in the networks of Fig.~\ref{f00099}.
The parameter $c_{opt}$ is the slope obtained by fitting a straight line to the peaks of the distributions of trained weights $w_f$ for fixed $w_i$. 
These results are averages measured in ten independent realizations, and the shaded areas represents the dispersion (standard deviation across realizations).
}%
\label{f01131}
\end{figure*}

\begin{figure*}[!tbh]
\centering
\includegraphics[width=\textwidth]{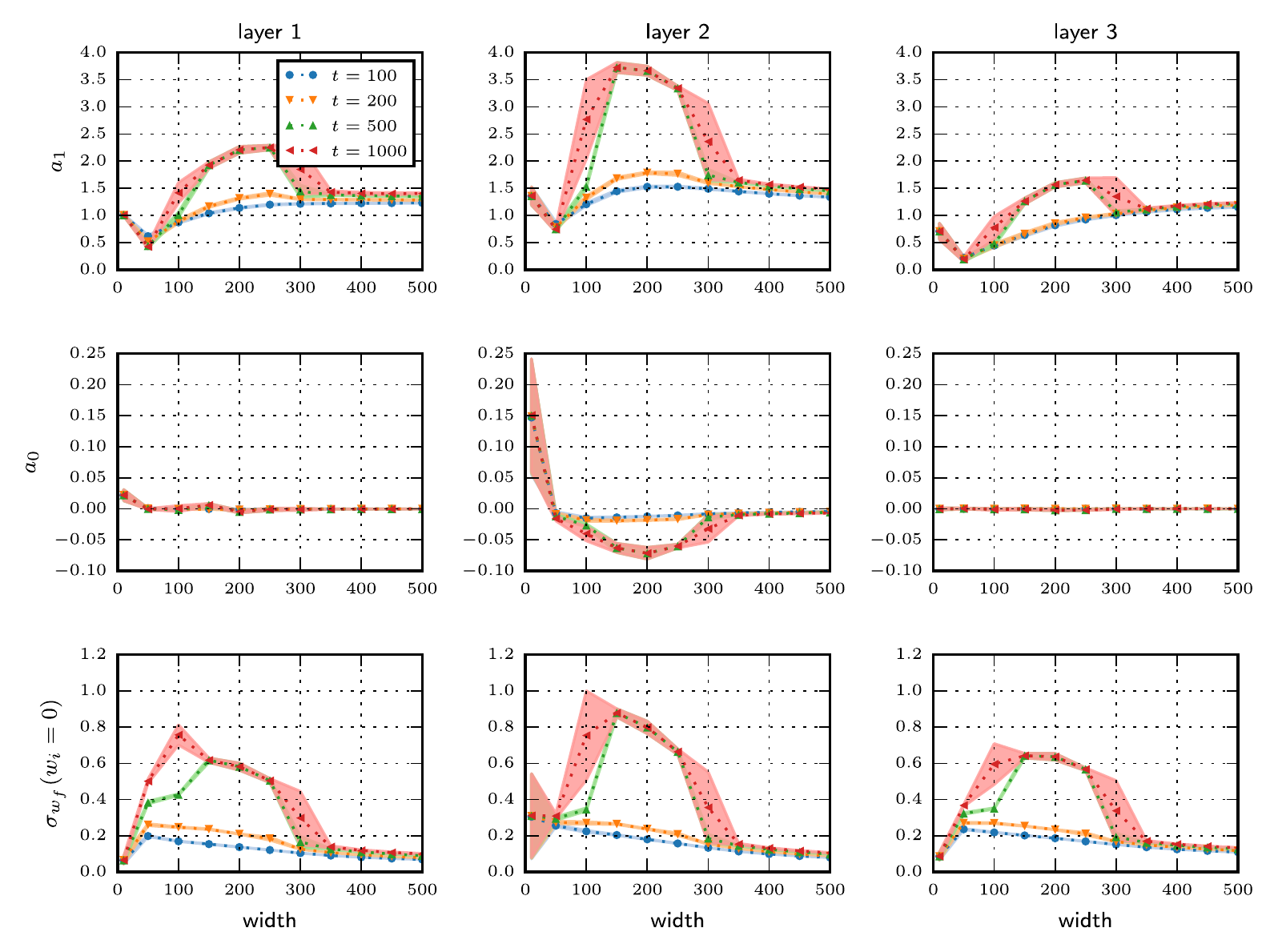}
\caption{Mean and standard deviation of the trained weights in the networks of Fig.~\ref{f00099}. 
Top and middle rows: results of fitting a straight line $a_0 + a_1 w_i$ to the mean of the trained weights $w_f(wi)$.
Bottom row: standard deviation of the distribution of trained weights for the set of weights that are initialized with the value $w_i=0$.
These results are averages measured in ten independent realizations, and the shaded areas represents the dispersion (standard deviation across realizations).
}%
\label{f01126}
\end{figure*}

The distribution of deviations from the initial weight is qualitatively similar across the trainable phase,
The proximity between the line $w_f=w_i$ and the median of the distributions of $w_f$ for fixed $w_i$, observed in Fig.~\ref{f50125}, is a distinctive feature of that kind of distribution.
Figure~\ref{f01131} shows the slope obtained by fitting a straight line to the peaks (or mode) of the distributions of trained weights $w_f$ for fixed $w_i$.
In large networks, the fitted slope of the peaks, $c_{opt}$, is very close $1$ in all layers (even for very large training times), independently of the width.
Below the width threshold for the network to be trainable, of about 300 nodes per hidden layer, the slope $c_{opt}$ shows significant deviations from $1$, and its value strongly fluctuates among realizations of the initialization and training. 
(The borders of the shaded area in the plots of Figs.~\ref{f01131} and~\ref{f01126} represent the standard deviation measured in ten independent realizations.) 
To emphasize the coupling between trainability and proximity to the initial configuration of weights, we used data from the same ten realizations to plot Fig.~\ref{f00099} and Figs.~\ref{f01131} and~\ref{f01126}. 
These figures combined show the simultaneousness of the abrupt increase of the loss and of the deviation from the initial configuration.

For the sake of completeness, we perform linear fittings also to the mean trained weight as a function of the initial weight.
Figure~\ref{f01126} shows the results of these fittings: $a_0$ and $a_1$ denote the constant and the slope, respectively.
Similarly to the $c_{opt}$, while above a width of about $300$ nodes the values of $a_0$ and $a_1$ are stable, below the threshold they suffer an abrupt change at some moment of training. 
The dispersion of the trained weights around their initial value, measured by the standard deviation, is also shown in Fig.~\ref{f01126} for the set of weights that are initialized with the value $w_i=0$, displaying the same transition at a width of about $300$. 
We observed that the distribution of trained weights for other $w_i \neq 0$ behaves similarly with the variation of the network's width.
Notice that, since the weights are initialized from a continuous distribution, we are able to measure the mode (peaks), average, and standard deviation of weights as functions of the initial value $w_i$ by applying the special procedure described in~\ref{appendixB}, which is less affected by the presence of noise (fluctuations) in the data that the standard binning methods.

\section{Fitting the statistics of weights in a single realization}
\label{appendixB}

\counterwithin{figure}{section}
\setcounter{figure}{0}
\renewcommand\thefigure{B.\arabic{figure}}  

This appendix briefly describes the methods used in this work to characterize the statistics 
of the displacements of weights produced by training with SGD.
The problem is that we cannot directly obtain the distributions $P_{w_i}(w_f)$ of the final weights $w_f$ 
for each value of the initial weight $w_i$,
because the $w_i$'s are drawn from a continuous (uniform) 
distribution, see Eq.~(\ref{e10}).
In practice, for a single realization of training, we have a have a set of points
 $(w_i, w_f)$, one for each link, in the continuous plane, as shown in Fig.~\ref{f50125}.
In this situation, calculating the mean $\langle w_f\rangle(w_i)$ and standard deviation $\sigma_{w_f}(w_i)$ 
of the distribution $P_{w_i}(w_f)$ may follow one
of two approaches: either using a binning procedure or cumulative distributions.
In our analysis, we employed the latter, which is less affected by random fluctuations 
than the standard binning methods.

We assume a linear fit $ \langle w_f\rangle(w_i) =   a_0 + a_1 w_i $, and obtain the
 constants $a_1$ and $a_0$ as follows. 
Let us define the function
\begin{equation}
W_f(w) = \int_{\min(w_i)}^w \langle w_f\rangle(x) dx = C + a_0 w + \frac{a_1}{2} w^2,
\end{equation}
where $C$ is a constant resulting from the lower limit of the integral.
For one
given realization, we can estimate this function from the following cumulative sum
\begin{equation}
    W_f(w) \approx \frac{\max(w_i)-\min(w_i)}{N} \sum_{j:w_j(0) \leq w} w_j(t_f),    
\label{eB2}
\end{equation}
where 
$w_j(t_f)$ denotes the value of the weight of link $j$ at time $t_f$,
$N$ is the number of links of the network, and $\min(w_i)$/$\max(w_i)$
is the minimum/maximum value of the initialization weights. 
[The sum in the right-hand side of Eq.~(\ref{eB2} runs over all links whose initial
 weight is not larger than $w$.]
Finally, we fit a second degree polynomial to $W_f(w)$, and 
get the constants $a_1$ and $a_0$ from its coefficients.

We use the same `cumulative-based' approach to find the second 
moment of $P_{w_i}(w_f)$, denoted by $\langle w_f^{2}\rangle (w_i) $.
In this case, we assume the polynomial
$\langle w_f^{2}\rangle (w_i) = b_0 + b_1 w_i + b_2 w_i^2$.
We again define the cumulative 
$W^{(2)}_f(w) = \int_{\min(w_i)}^w \langle w_f^2 \rangle(x) dx $. 
Similarly to $W_f(w)$, we estimate $W^{(2)}_f(w)$ as
\begin{equation}
    W^{(2)}_f(w) \approx \frac{\max(w_i)-\min(w_i)}{N}  \sum_{j:w_j(0) \leq w}
    {w_j}^2(t_f),
    \label{eB4}
\end{equation}
and fit a third degree polynomial to get the coefficients $b_0$, $b_1$, and $b_2$.
Then, we calculate $\sigma_{w_f}(w_i)$ as
\begin{equation}
    \sigma_{w_f}(w_i) = \sqrt{\langle w_f^{2} \rangle(w_i) -
    {\left[\langle w_f \rangle(w_i)\right]}^2}.
    \label{eB3}
\end{equation}

The method for fitting the peak (or the mode) of the distribution $P_{w_i}(w_f)$ is 
also based on a cumulative distribution.
In our experiments we observe that, in the trainability regime, the peak of the
distribution of $w_f$ as a function of $w_i$ is indistinguishable from a 
straight line, see Fig.~\ref{f50125}.
Accordingly, we define 
\begin{equation}
    N_{c}(b) = \left| \left\{(w_i, w_f)\colon w_f \le b + c w_i\right\} \right|,
\end{equation}
which is a function that counts the number of points $(w_i,w_f)$ below or at the line $b+c w_i$.
Then, we fit the peak of $P_{w_i}(w_f)$  by optimizing the expression
\begin{equation}
    \max_{c} \max\left(\frac{d}{db}N_{c}(b)\right),
\end{equation}
In other words, we look for the slope that causes
the largest rate of change in the function $N_c(b)$.
This slope, $c_{opt}$, is
the slope of the linear function that best aligns with the peak of the
distribution $P_{w_i}(w_f)$.

\bibliography{main}

\end{document}